\documentclass{article}

\usepackage{microtype}
\usepackage{graphicx}
\usepackage{subfigure}
\usepackage{booktabs} % for professional tables
\usepackage{multirow}
\usepackage[normalem]{ulem}
\useunder{\uline}{\ul}{}
\usepackage{adjustbox}

\usepackage{hyperref}

\usepackage{abbrv}

\usepackage[accepted]{icml2025}

\usepackage{amsmath}
\usepackage{amssymb}
\usepackage{mathtools}
\usepackage{amsthm}

\usepackage[capitalize,noabbrev]{cleveref}

\theoremstyle{plain}
\newtheorem{theorem}{Theorem}[section]

\theoremstyle{definition}
\newtheorem{definition}[theorem]{Definition}

\theoremstyle{remark}

\usepackage[textsize=tiny]{todonotes}

\usepackage[font=small,labelfont=bf,tableposition=top]{caption}
\usepackage{cuted} % For the strip environment
\icmltitlerunning{Separating Knowledge and Perception with Procedural Data}

\begin{document}

\twocolumn[
\icmltitle{Separating Knowledge and Perception with Procedural Data}




\begin{icmlauthorlist}
\icmlauthor{Adrián Rodríguez-Muñoz}{mit}
\icmlauthor{Manel Baradad}{mit}
\icmlauthor{Phillip Isola}{mit}
\icmlauthor{Antonio Torralba}{mit}
\end{icmlauthorlist}

\icmlaffiliation{mit}{Department of Electrical Engineering and Computer Science, Massachusetts Institute of Technology, Cambridge, USA}
%\icmlaffiliation{comp}{Company Name, Location, Country}
%\icmlaffiliation{sch}{School of ZZZ, Institute of WWW, Location, Country}

\icmlcorrespondingauthor{Adrian Rodríguez-Muñoz}{adrianrm@mit.edu}
%\icmlcorrespondingauthor{Firstname2 Lastname2}{first2.last2@www.uk}

\icmlkeywords{Machine Learning, ICML}

\vskip 0.3in
]



\printAffiliationsAndNotice{}  % leave blank if no need to mention equal contribution

%\begin{center}
%    \includegraphics[width=\linewidth]{figures/fig1.pdf}
%    \captionof{figure}{While solid paints (left top).}\label{fig:teaser}
%\end{center}

\begin{strip}
    \vspace{-2.4cm}
    \centering
    \includegraphics[width=\textwidth]{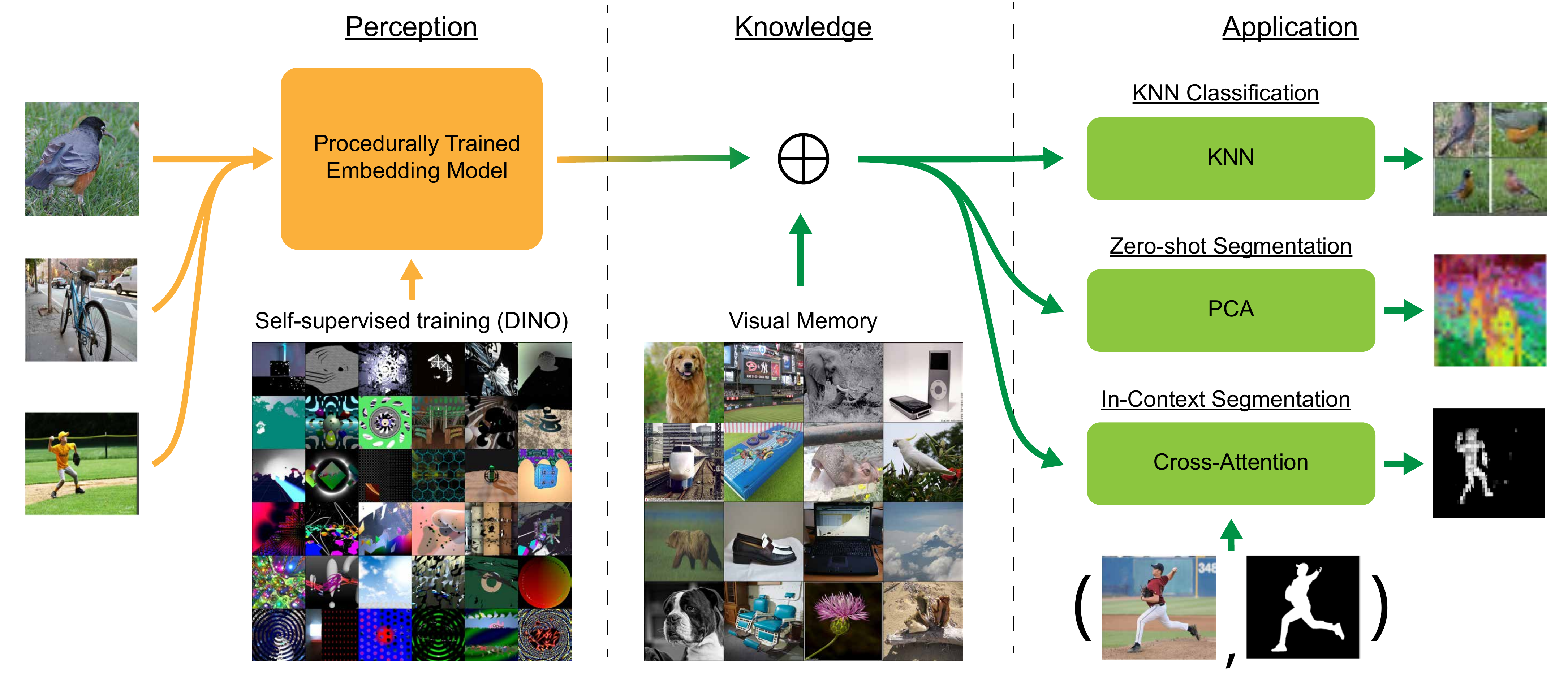}
    \captionof{figure}{Our approach is as follows: first, an embedding model is trained on procedural data generated with OpenGL code using self-supervised learning (SSL). In this stage, we learn useful visual features while constraining real-world knowledge entering the model. Next, we introduce real-world knowledge into the model using only a visual memory of reference image embeddings, without extra training. Lastly, we apply the memory-augmented model on visual similarity, classification, and segmentation tasks. The overall system has perfect control over all real world data, while approximating the performance of standard training. Moreover, isolating all real data to only the memory makes efficient data unlearning and privacy analysis possible.}
    \label{fig:teaser}
\end{strip}

\begin{abstract}
% What we do
We train representation models with procedural data only, and apply them on visual similarity, classification, and semantic segmentation tasks without further training by using visual memory---an explicit database of reference image embeddings.
% Contributions
Unlike prior work on visual memory, our approach achieves full compartmentalization with respect to all real-world images while retaining strong performance.
% Performance
Compared to a model trained on Places, our procedural model performs within 1\% on NIGHTS visual similarity, outperforms by 8\% and 15\% on CUB200 and Flowers102 fine-grained classification, and is within 10\% on  ImageNet-1K classification. It also demonstrates strong zero-shot segmentation, achieving an $R^2$ on COCO within 10\% of the models trained on real data.
% Analysis
Finally, we analyze procedural versus real data models, showing that parts of the same object have dissimilar representations in procedural models, resulting in incorrect searches in memory and explaining the remaining performance gap.
\end{abstract}

%\clearpage
\begin{figure*}[t]
    \centering
    \includegraphics[width=\linewidth, height=\textheight, keepaspectratio=True]{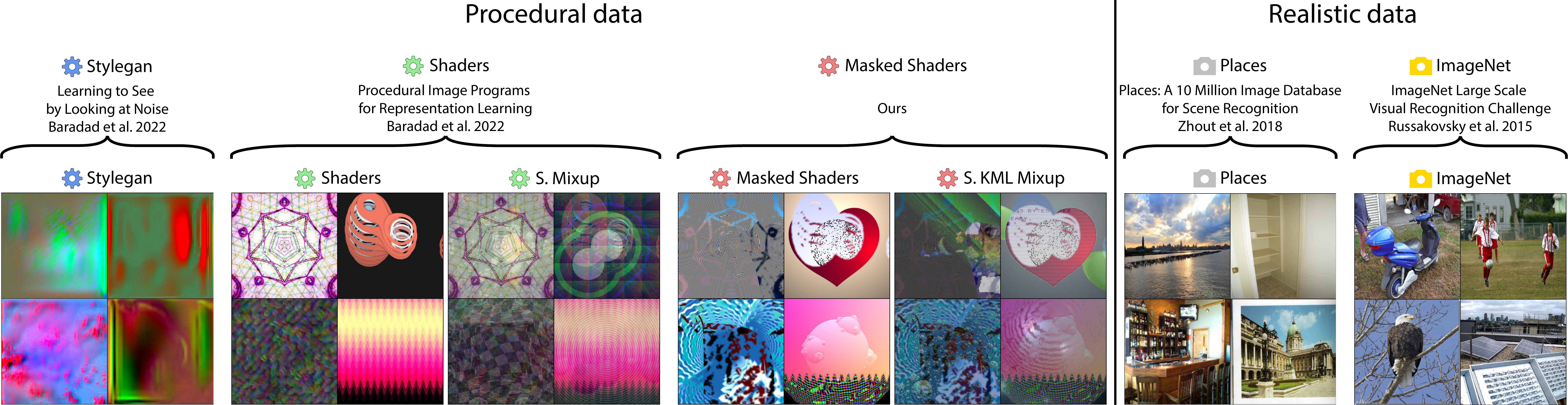}
    \captionof{figure}{Examples of procedural data from prior work, our new Masked Shaders: Shaders KML and Shaders KML Mixup, and the real datasets Places and ImageNet. Masked Shaders have higher diversity and consistently beat prior processes in downstream tasks.}
    \label{fig:datasets-examples}
    \vspace{-0.2cm}
\end{figure*}
\section{Introduction}

Modern vision systems learn by digesting images into weights via gradient descent. While offering strong performance, this approach carries concerns about interpretability, privacy, and bias. Moreover, it makes adding and removing data difficult, since weights store knowledge in a black-box manner. To alleviate this issue, prior work proposed the idea of explicitly separating how knowledge is represented, the feature embeddings, from what knowledge is stored, the visual memory \cite{geirhos_towards_2024, weston_memory_2015, chen_semi-supervised_2018, iscen_memory_2022, iscen_retrieval-enhanced_2024, gui_knn-clip_2024, silva_learning_2024}. They called this approach \textbf{``perception with visual memory"}.

Unlike a traditional network which feeds the query embedding through a classifier to obtain the output, perception with visual memory works by taking the query embedding, retrieving its k-nearest neigbhours (KNN) from the memory, and outputting the majority label. This in turn makes adding and removing knowledge easy and efficient: while the classifier would need to be fine-tuned or retrained, the visual memory can simply add and drop data samples. There remains just one problem, the feature embeddings are also a parametric model trained on data. Thus, knowledge and perception are not fully separated. Moreover, while adding and removing training samples from the memory is easy, doing so from the feature embeddings is not. In this work, we propose training the embedding model with procedural data. Unlike real world data, procedural data is non-realistic and is generated via simple code, which allows us to achieve a stronger separation between perception and knowledge as seen in \cref{fig:teaser}. \cref{fig:datasets-examples} compares examples of procedural and real images. Prior work focused on combining procedural embeddings with linear classifiers trained on real data \cite{baradad_learning_2022, baradad_procedural_2023}, and only very briefly mentioned using a neighbors approach. In this work, we delve deeper into this ability and explicitly make the connection to visual memory perception.

Our contributions are as follows:
\begin{enumerate}
    \item We demonstrate that procedural embeddings with visual memory allow perfect unlearning and privacy guarantees \wrt \emph{all} real data, while retaining strong performance.
    \item We introduce the new procedural data processes Shaders KML and Shaders KML Mixup, which yield stronger embeddings than those of prior work.
    \item We show that procedural embeddings possess remarkable zero-shot and in-context semantic segmentation abilities, on the same order of magnitude as embeddings trained on real data.
\end{enumerate}

\section{Related work}

\textbf{Visual memory} marries the compartmentalization and interpretability of databases with the effectiveness of neural methods. This is done by applying k-nearest neighbours algorithms on trained neural embedding databases \cite{papernot_deep_2018}. Prior work has proposed employing this technique for few-shot learning \cite{wang_simpleshot_2019, yang_d2n4_2020, bari_nearest_2021}, adversarial robustness \cite{sitawarin_robustness_2019, papernot_deep_2018, rajani_explaining_2020}, medical image classification
\cite{martel_deep_2020}, confidence calibration \cite{papernot_deep_2018}, interpretability \cite{papernot_deep_2018, rajani_explaining_2020, wallace_interpreting_2018, lee_improving_2020}, image denoising \cite{plotz_neural_2018}, retrieval-
augmented learning \cite{khandelwal_generalization_2020, drozdov_you_2022}, anomaly and out-of-distribution detection \cite{bergman_deep_2020, sun_out--distribution_2022}, and language models \cite{wu_memorizing_2022, khandelwal_generalization_2020, min_silo_2024}. In our work, we specifically build on top of \cite{nakata_revisiting_2022} and \cite{geirhos_towards_2024}, which showed the effectiveness of visual memory at large scale. Different from them, we use procedural data to train the embeddings, thus making \emph{all} real data used be under database-like control.

\textbf{Procedural data} approaches train networks on non-realistic data generated via code. The benefits of this are several fold: it makes the process more interpretable and optimizable \cite{sun_autoflow_2021}, can be used to improve the privacy-utility tradeoff of differentially private gradient descent (SGD) \cite{abadi_deep_2016, tang_differentially_2023, choi_leveraging_2024}, and provides insights on the human visual system. Prior work has explored using fractals \cite{kataoka_pre-training_2021, anderson_improving_2021, nakashima_can_2021}, untrained generative networks \cite{baradad_learning_2022}, and recently, openGL programs \cite{baradad_procedural_2023}. We contribute two new processes that obtain higher performance, and show the semantic segmentation ability of procedural models.

%\clearpage
\section{Train procedural embeddings}
\label{sec:stage-1}
We train an embedding model using \emph{procedural} data, similarly to \cite{baradad_learning_2022, baradad_procedural_2023}. Unlike typical synthetic data which approximates the target distribution using a complex generative model, procedural data is created with simple programs. See \cref{fig:datasets-examples} for the procedural and real datasets considered in this work. We train vision transformers (ViT-S) \cite{dosovitskiy_image_2021} using the local-to-global similarity objective of DINO \cite{caron_emerging_2021}. On real data, this objective teaches models to have similar representations for parts of objects existing in reality, leaking biases and personal identities in the process. In contrast, the same objective but with procedural data teaches to have similar representations for parts of abstract shapes and textures, with much less risk of biases and privacy leakage. Procedural embeddings lack knowledge of real world entities, yet are surprisingly strong.

\begin{figure}
    \centering
    \includegraphics[width=\linewidth]{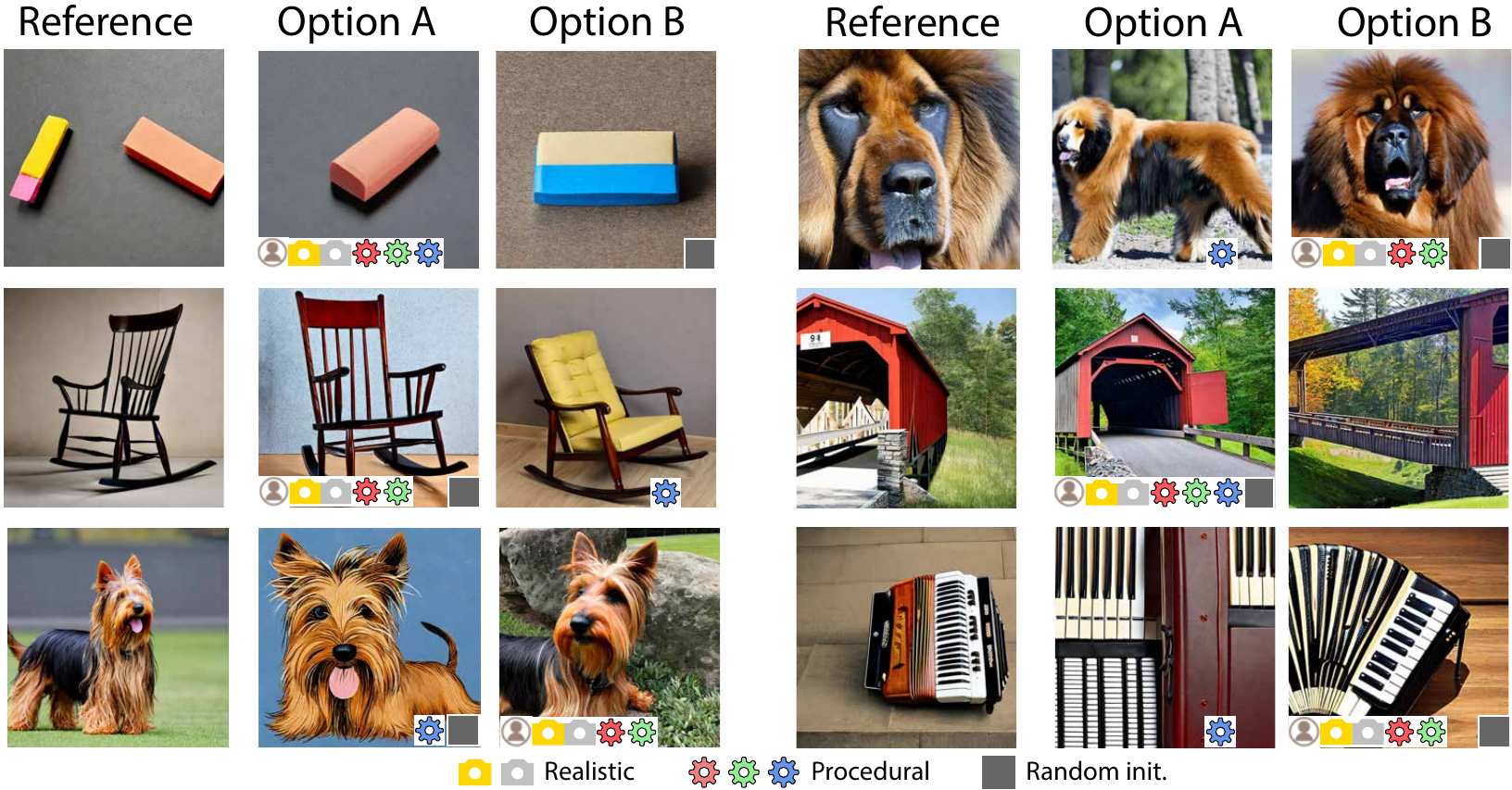}
    \caption{Examples from the NIGHTS dataset, along with human, real data model, and procedural model judgments.}
    \label{fig:nights-hvs-examples}
    %\vspace{-0.2cm}
\end{figure}

\begin{figure}
    \centering
    \includegraphics[width=\linewidth]{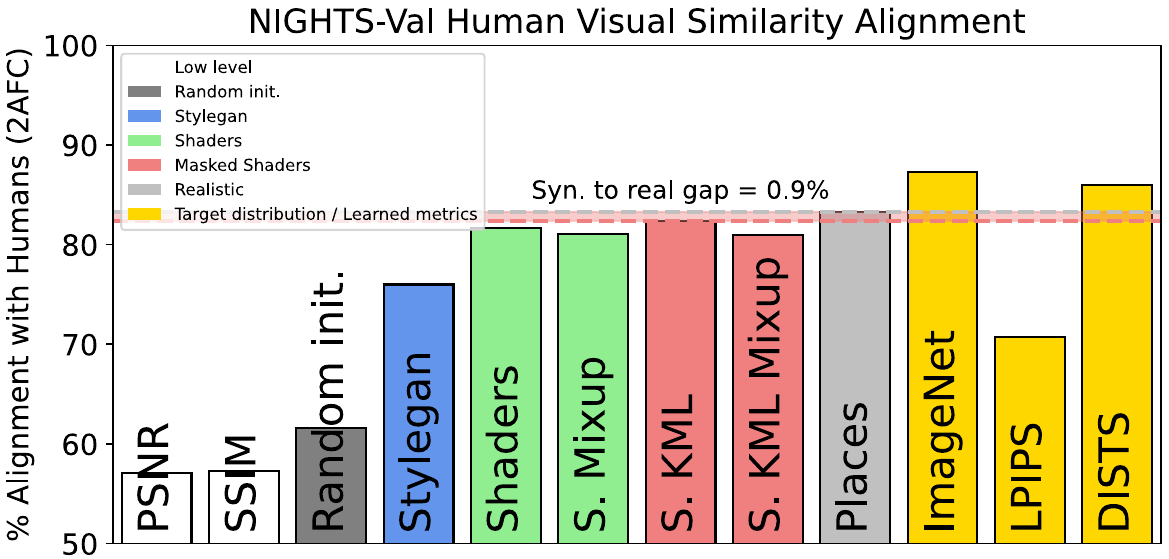}
    \caption{Models performance on the NIGHTS-Val benchmark. The best procedural model, trained on Shaders KML, has a high \% alignment with humans of 82.4\%, within 0.9\% of the Places model, trained on real data without domain overlap.}
    \label{fig:nights-hvs-perf}
    %\vspace{-0.5cm}
\end{figure}

We evaluate the procedural networks on a Human Visual Similarity (HVS) task using the NIGHTS dataset \cite{fu_dreamsim_2023}, an analysis missing in prior work. This benchmark consists of a Two Alternative Forced Choice (2AFC) on trios of images: given a reference and two options, which option has greater embedding cosine similarity with the reference? The test measures ability to match human judgments on not only low-level colors and textures, which are easily captured by simple white-box metrics, but also on mid-level similarities in layout, pose, and content, that are harder to define. \cref{fig:nights-hvs-examples} shows examples from the dataset with human and model answers. Results in \cref{fig:nights-hvs-perf} show that procedural data metrics have performance within 1\% of the Places model, trained on real data without domain overlap. The ImageNet model has class overlap with NIGHTS, and thus is only for reference. White-box metrics like PSNR and SSIM barely perform above chance.

%\clearpage
%\vspace{-0.2cm}
\subsection{The Shaders KML Mixup process}
The prior best procedural dataset is Shaders Mixup from \cite{baradad_procedural_2023}. They observed that models trained on the raw Shaders dataset learned short-cut solutions with poor generalization, but interpolating multiple samples in pixel space with Mixup \cite{zhang_mixup_2018} alleviated the issue and achieved a new state-of-the-art.
\begin{figure}
    \centering
    \includegraphics[width=\linewidth, height=\textheight, keepaspectratio=True]{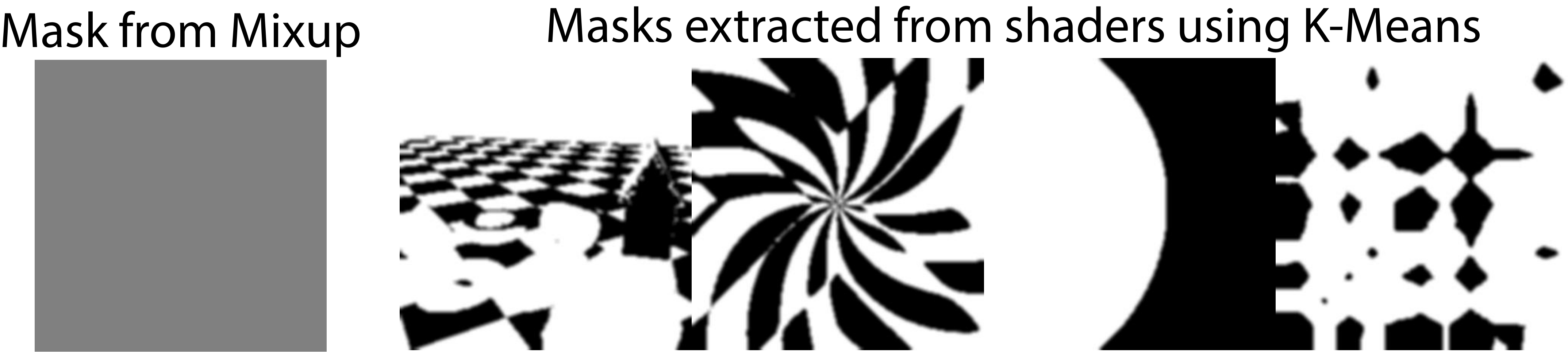}
    \caption{Constant mixing mask used by mixup (left) vs data driven mixing masks obtained using KMeans (right). Using the latter leads to much greater diversity.}
    \label{fig:mixing-masks}
    %\vspace{-0.3cm}
\end{figure}

\begin{figure}
    \centering
    \includegraphics[width=\linewidth, height=\textheight, keepaspectratio=True]{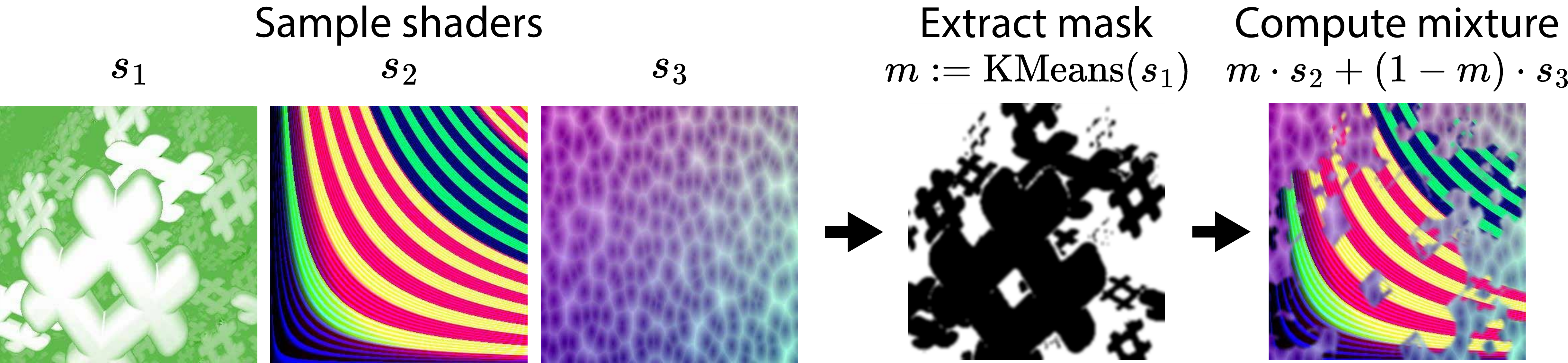}
    \caption{Diagram of the Shaders KML process. First, three shaders $s_1$, $s_2$, and $s_3$ are sampled. Next, $s_1$ is used to obtain a mask $m$ using KMeans in RGB space. Finally, we mix $s_2$ and $s_3$ using $m$ to obtain the Shaders KML sample.}
    \label{fig:kml_process}
    %\vspace{-0.55cm}
\end{figure}

In our work, we derive a stronger approach that extracts mixing masks from the Shaders images, rather than always using a constant mixing mask like in Mixup. As seen in \cref{fig:mixing-masks} this has the effect of increasing dataset diversity, found in \cite{baradad_learning_2022} to be one of the biggest drivers of performance for non-real data. We call this process, shown visually in \cref{fig:kml_process}, Shaders K-Means Leaves (S. KML). Shaders KML obtains comparable performance to Shaders Mixup. Applying Mixup on Shaders KML to supress short-cut solutions yields the Shaders KML Mixup process, obtaining a new state-of-the-art as seen in \cref{tab:classification}.
%\vspace{-0.3cm}
\section{Classification and segmentation with visual memory}\vspace{-0.1cm}
\label{sec:stage-2}
We apply the procedural models on classification and segmentation tasks without further training using visual memory. We compare procedural data with training on Places, a dataset of natural images different to the evaluation distribution, which acts as an upper bound to procedural data. Results for training on ImageNet, which is either the target dataset or has high overlap, is included only for reference.

\begin{figure}
    \centering
    \includegraphics[width=\linewidth, height=\textheight, keepaspectratio=True]{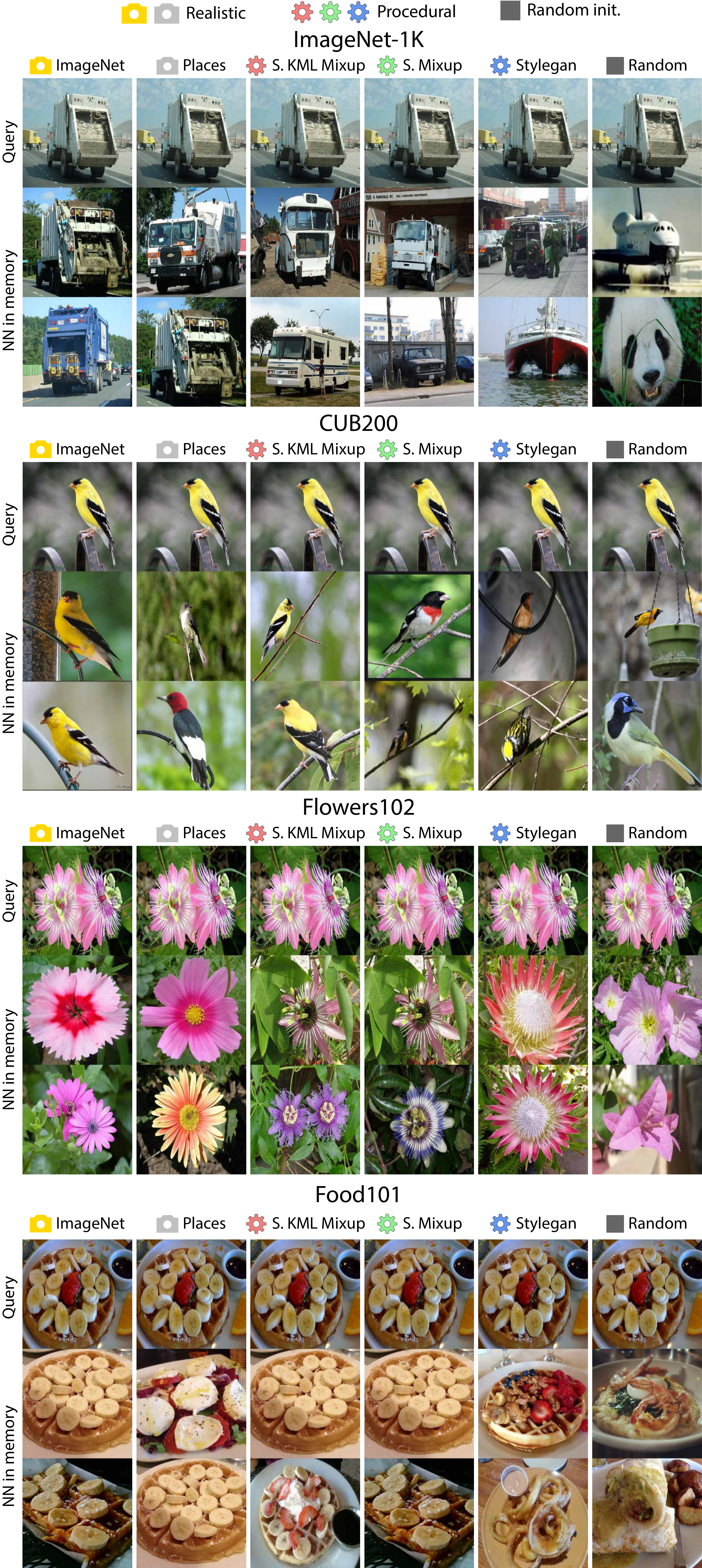}
    \caption{Visual comparison of query images from various datasets and their neareast neighbours according to each of the models. Procedural models can effectively search for perceptually similar images on a wide variety of datasets, despite not seeing real-world data during training.}
    \label{fig:neighbours_all}
    \vspace{-0.5cm}
\end{figure}

\textbf{Classification} is qualitatively more challenging than the similarity task from \cref{sec:stage-1}: going from mid-level concepts in a small three image set, to higher-level semantics in a large and diverse look-up pool of up to $\mathcal{O}$(1M) images. However, procedural models still obtain strong performance. \cref{tab:classification} shows KNN classification accuracy on various fine-grained datasets and ImageNet-1K \cite{russakovsky_imagenet_2015}. Remarkably, the best procedural model actually beats the model trained on Places \cite{zhou_places_2018} on the fine-grained classification datasets, by 15\%, 8\%, and 1\% on Flowers102 \cite{nilsback_automated_2008}, CUB200 \cite{wah_caltech-ucsd_2011}, and Food101 \cite{fleet_food-101_2014} respectively. We posit that this is due to the little or no semantic overlap between Places and the three fine-grained datasets, which means the Places model spends capacity on semantic associations that are not useful. In contrast, the procedural models learn agnostic skills from abstract shapes and textures that may better generalize. As a sanity check, all procedural models are worse than the ImageNet model which has semantic overlap with all three fine-grained datasets. On ImageNet-1K classification, the best procedural model is within $<$\hspace{-.01cm}10\% of the Places model. \cref{fig:neighbours_all} shows a query image and its nearest neighbors (NNs) for three datasets, visually showing the abilities of procedural models.
\begin{table}
    %\vspace{-3.5cm}
    \centering
    \begin{adjustbox}{max width=\linewidth, max height=\textheight}
% \usepackage{booktabs}
% \usepackage{multirow}
% \usepackage[normalem]{ulem}
\begin{tabular}{@{}ll|rrr|r@{}}
\multicolumn{1}{c}{Data type} & \multicolumn{1}{c|}{Dataset} & \multicolumn{3}{c|}{Fine-grained}                                                 & \multicolumn{1}{c}{General}    \\
\multicolumn{1}{c}{}             & \multicolumn{1}{c|}{}        & \multicolumn{1}{c}{Flowers} & \multicolumn{1}{c}{CUB} & \multicolumn{1}{c|}{Food} & \multicolumn{1}{c}{ImageNet-1K} \\ \midrule
Target                          & ImageNet                     & 83.43                       & 55.20                   & 64.45                     & 68.89                           \\ \midrule
Realistic                        & Places                       & {\ul 59.51}                 & {\ul 19.09}             & {\ul 47.78}               & {\ul 47.30}                     \\ \midrule
\multirow{5}{*}{Procedural}      & S. KML Mixup                 & \textbf{75.20}              & \textbf{27.08}          & \textbf{48.70}            & \textbf{37.88}                  \\
                                 & S. KML                       & 71.86                       & 24.02                   & 46.00                     & 35.38                  \\
                                 & S. Mixup                     & 73.33                       & 19.85                   & 47.78                     & 35.45                           \\
                                 & Shaders                      & 66.18                       & 15.59                   & 38.05                     & 30.69                           \\
                                 & Stylegan                     & 41.86                       & 8.61                    & 22.73                     & 13.73                           \\ \midrule
\multirow{2}{*}{White-box}       & Random init.                 & 11.18                       & 1.93                    & 5.32                      & 1.84                            \\
                                 & SIFT                         & -                           & -                       & -                         & 3.08                           
\end{tabular}
    \end{adjustbox}
    \caption{Performance on KNN classification with visual memory. The best procedural model, trained on Shaders KML Mixup, beats the real data Places model on Flowers, CUB, and Food fine-grained classification, and has a gap of only 10\% on ImageNet-1K.}
    \label{tab:classification}
\end{table}

\clearpage
\begin{strip}
    \centering
    \includegraphics[width=\linewidth, height=\textheight, keepaspectratio=True]{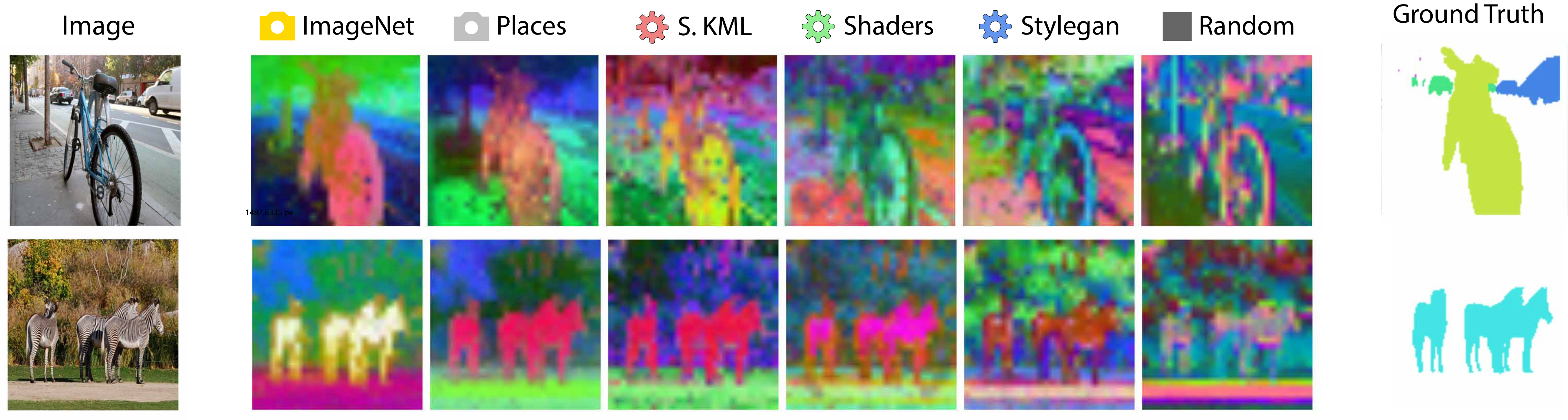}
    \captionof{figure}{Zero-shot segmentation on COCO \cite{lin_microsoft_2015} using principal component analysis (PCA) features. Procedural models clearly separate the bike and zebra from the background. However, visually distinct parts of the bike, such as the center and spokes of the wheel, have similar and dissimilar representations in real and procedural models respectively. This is due to procedural models not having seen bikes before, while real models learn they are parts of the same object.}
    \label{fig:segmentation-zeroshot}
    \vspace{0.1cm}

    \centering
    \includegraphics[width=\linewidth, height=\textheight, keepaspectratio=True]{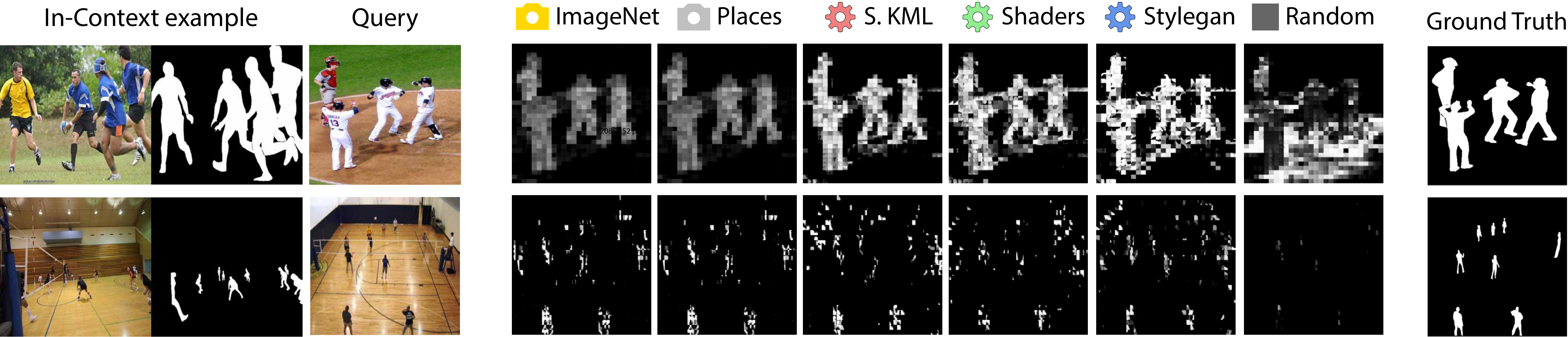}
    \captionof{figure}{In-context segmentation on Ade20k \cite{zhou_semantic_2018}. Procedural models can segment arbitrary classes given a single exemplar prompt.}
    \label{fig:segmentation-incontext}
    %\vspace{0.4cm}
\end{strip}
\begin{figure}
    \centering
    \includegraphics[width=\linewidth, height=.22\textheight, keepaspectratio=True]{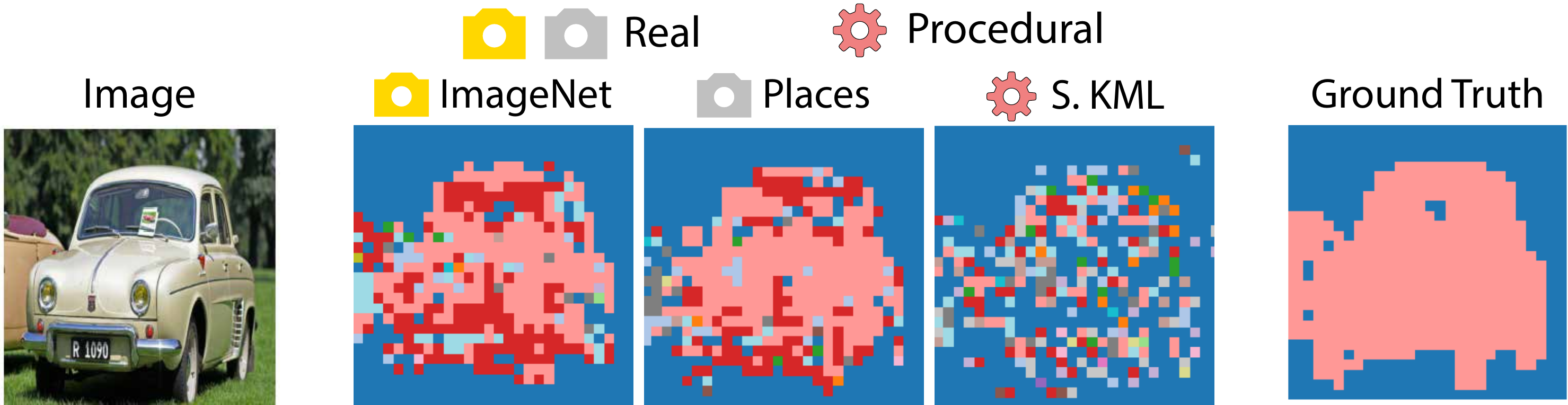}
    \captionof{figure}{KNN segmentation on Pascal \cite{everingham_pascal_2010}. Procedural models are limited at segmenting with KNN. Due to not seeing real-world objects during training, representations of individual parts can be very dissimilar. This leads to spurious similarities with object parts of other classes, harming performance.}
    \label{fig:segmentation-knn}
    %\vspace{-0.7cm}
\end{figure}
\textbf{Segmentation:} Procedural models have remarkable semantic segmentation ability. \cref{fig:segmentation-zeroshot} qualitatively shows how procedural features clearly separate the bike and zebras from their surroundings. Quantitatively, \cref{tab:segmentation-r2} shows numerical $R^2$ (ratio of explained variance to total variance) between principal component analysis (PCA) features and human labels. The best procedural model is within 10\% of real data models and highly above random and RGB features. Procedural models are also capable of in-context segmentation: given a prompt image and a prompt mask representing a concept, they can effectively search for it in a new query image, as shown in \cref{fig:segmentation-incontext}. However, they struggle at KNN semantic segmentation with a large visual memory, as seen in \cref{fig:segmentation-knn}. As explained in \cref{sec:stage-1}, the DINO objective on real data teaches models to have similar representations for parts of real world objects, even when the parts are visually dissimilar. In contrast, procedural models, having never seen the object during training, will have dissimilar representations for the parts. We can observe this in \cref{fig:segmentation-zeroshot}: the center and spokes of the wheel are colored the same in real models and differently in procedural models. Procedural models have excessively local representations which are vulnerable to spurious similarities with object parts of other classes.
\begin{table}
    %\vspace{-2cm}
    \centering
    \begin{adjustbox}{max width=\linewidth, max height=.15\textheight}
% \usepackage{booktabs}
% \usepackage{multirow}
% \usepackage[normalem]{ulem}
\begin{tabular}{@{}llr@{}}
\multicolumn{1}{c}{Data type} & \multicolumn{1}{c|}{Dataset}      & \multicolumn{1}{c}{COCO} \\ \midrule
\multirow{2}{*}{Realistic}       & \multicolumn{1}{l|}{ImageNet}     & {\ul 63.7}               \\ \cmidrule(l){2-3} 
                                 & \multicolumn{1}{l|}{Places}       & 62.1                     \\ \midrule
\multirow{5}{*}{Procedural}      & \multicolumn{1}{l|}{S. KML Mixup} & 53.7                     \\
                                 & \multicolumn{1}{l|}{S. KML}       & \textbf{55.9}            \\
                                 & \multicolumn{1}{l|}{S. Mixup}     & 51.4                     \\
                                 & \multicolumn{1}{l|}{Shaders}      & 55.0                     \\
                                 & \multicolumn{1}{l|}{Stylegan}     & 48.5                     \\ \midrule
\multirow{2}{*}{White-box}       & \multicolumn{1}{l|}{Random init.} & 36.7                     \\
                                 & RGB                               & 19.4                    
\end{tabular}
    \end{adjustbox}
    \caption{$R^2$ of PCA features and human label segmentations.}
    \label{tab:segmentation-r2}
    \vspace{-0.4cm}
\end{table}

% Limitations
\clearpage
\begin{strip}
    \centering
    \includegraphics[width=\linewidth, height=.67\textheight, keepaspectratio=True]{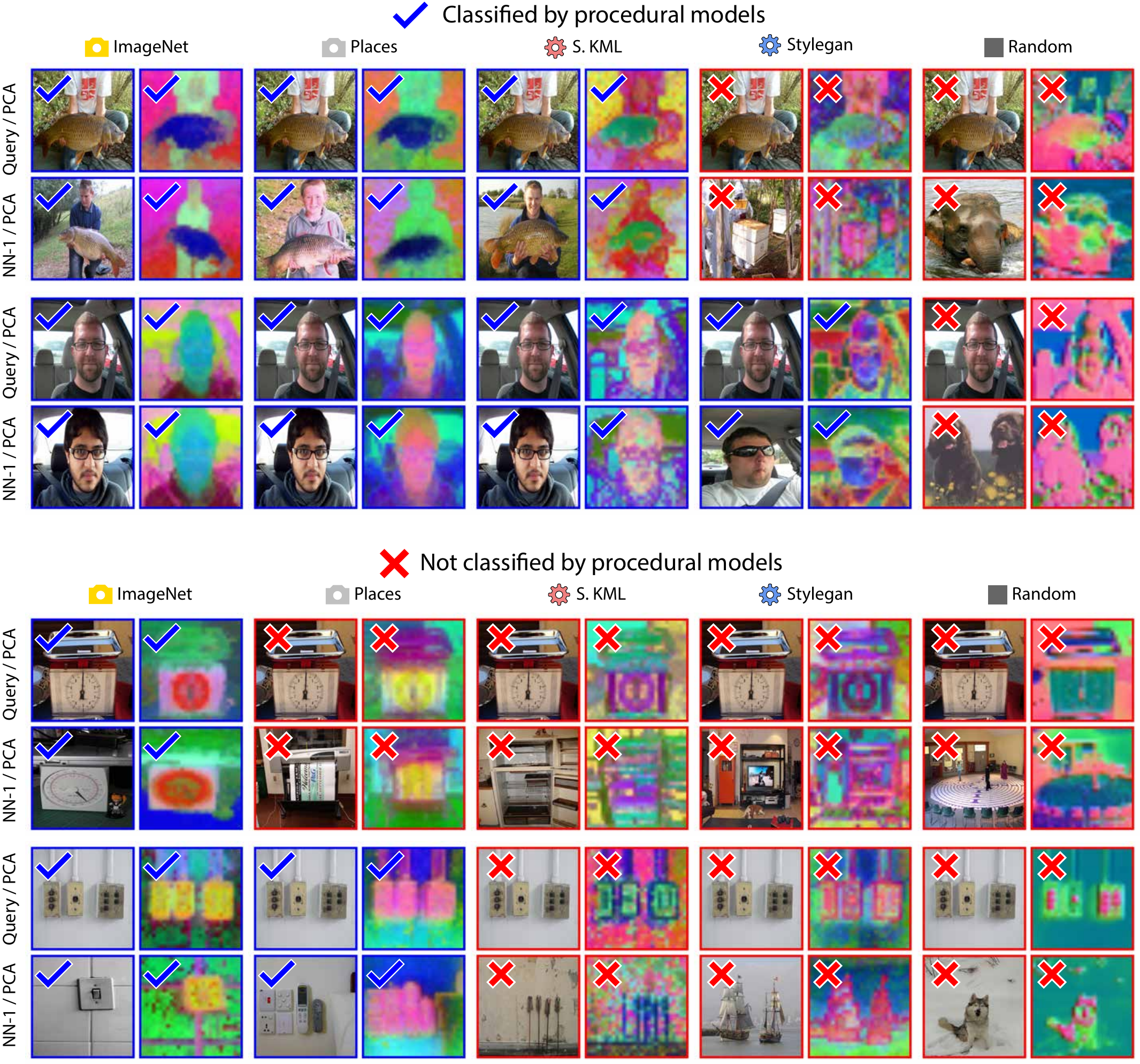}
    \captionof{figure}{Feature PCAs of images correctly (top) and incorrectly (bottom) classified by the Shaders KML Mixup model. PCAs for correctly classified images separate distinct objects and join parts of the same object (fish, face). In the other hand, in incorrectly classified images they fail to separate distinct objects (wall and body of the plugs) or fail to join parts of the same object (hand and face of the dial).}
    \label{fig:limitations-pca}
\end{strip}

\section{Analysis of limitations}
\label{sec:limitations}

Despite their good performance, procedural models still lag behind models trained on real data. In this section, we develop insights on why and how this gap occurs. \cref{fig:limitations-pca} shows feature PCAs of images and their nearest neighbor in memory according to each model. We observe that correctly classified images have feature PCAs close to the natural segmentation: distinct objects are clearly separated, and parts of the same object such as the fish and the face share a single distinct color. In contrast, for images classified incorrectly PCAs fail to separate distinct objects (the switch cable and casing have very similar colors) or fail to join parts of the same object (the hand and face of the dial have different colors), problems which are not the case in the feature embeddings trained on ImageNet.

Due to never having seen them during training, procedural models do not know to identify the object that defines the class (e.g. dial, switch) as a single entity, leading to incorrect nearest neighbours. For example, in the balance image (first incorrect example), the S. KML Mixup model separates the metallic balance, dial face, and dial hand in green, yellow, and purple respectively. The nearest neighbour \emph{correctly} has similar looking regions, but is the \emph{wrong} class. We leave finding ways of addressing these limitations to future work.

\section{Efficient unlearning and privacy guarantees}
\label{sec:stage-3}

Lastly, we explore how our approach can efficiently unlearn data,  compute privacy guarantees, and handle sensitive data, problems that are expensive and difficult for prior approaches.

\textbf{Data unlearning} is the problem of eliminating a piece of data from the weights of a model. This is an important issue especially when it comes to images of humans or NSFW or illegal content, and has received widespread attention in generative models. Prior work has focused on editing the weights to suppress the concept, but current methods are not infallible and still use the "contaminated" weights as an initial point, which wouldn't satisfy a legal request. Visual memory \cite{geirhos_towards_2024} offers a compelling solution: simply remove the offending data from the memory. However, this approach fails when the offending data was used to train the embeddings. Training on procedural data solves this problem elegantly by suppressing this possibility, as procedural data has little privacy risk.
\begin{figure}[t]
    \centering
    \includegraphics[width=\linewidth, height=\textheight, keepaspectratio=True]{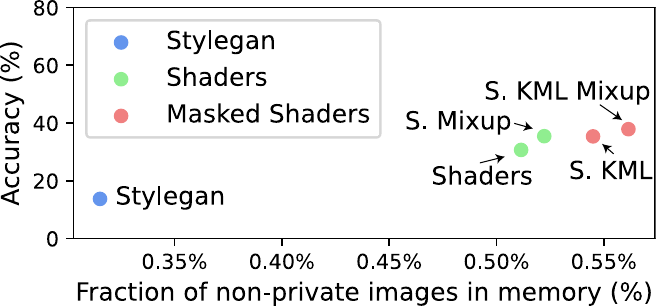}
    \caption{Non-private training samples (\%) versus KNN classification accuracy on ImageNet for the procedurally trained models. A linear relationship between accuracy and privacy is observed, though only 0.6\% of training samples are non-private.}
    \label{fig:privacy}
    %\vspace{-0.3cm}
\end{figure}

\textbf{Differential privacy} characterizes anonymity of individual samples in a training set \cite{dwork_algorithmic_2013, dwork_calibrating_2006, dwork_firm_2011}. Formally
\begin{definition}[$\epsilon$-differentially private algorithm]
    Let $\mathcal{A}$ be an algorithm that takes a dataset as input, and let $\text{Im }\mathcal{A}$ be the image of $\mathcal{A}$. Then $\mathcal{A}$ is $\epsilon$-differentially private with respect to sample $x$ if for all $D_1, D_2$ datasets differing only in $x$ and all subsets $S$ of $\text{Im }\mathcal{A}$, we have
    \begin{equation}
        \Pr[\mathcal{A}(D_1) \in S] \leq e^{\epsilon}\Pr[\mathcal{A}(D_2) \in S]
    \end{equation}
\end{definition}
Essentially, a sample $x$ is private if the distribution of the outputs of $\mathcal{A}$ trained with and without $x$ are close. For a deterministic algorithm, the definition simplifies greatly to the following: we have differential privacy \wrt $x$ if all predictions on the test set are the same when using and not using $x$. This is expensive to compute when $x$ was used to train the embeddings---the latter change, so re-training is required. In contrast, for procedural embeddings with visual memory we can simply compare test set predictions with and without any real image $x$ in the memory, which takes little time to compute. \cref{fig:privacy} plots KNN classification accuracy vs the fraction of non-private training images (those which, when removed from the memory, change the prediction of at least one test image) on ImageNet-1K. It shows a linear trend between performance and privacy, although the models are quite private as less than 0.6\% of the samples are non-private.

\begin{figure*} %[t]
    \centering
    \includegraphics[width=\linewidth, height=0.5\textheight, keepaspectratio=True]{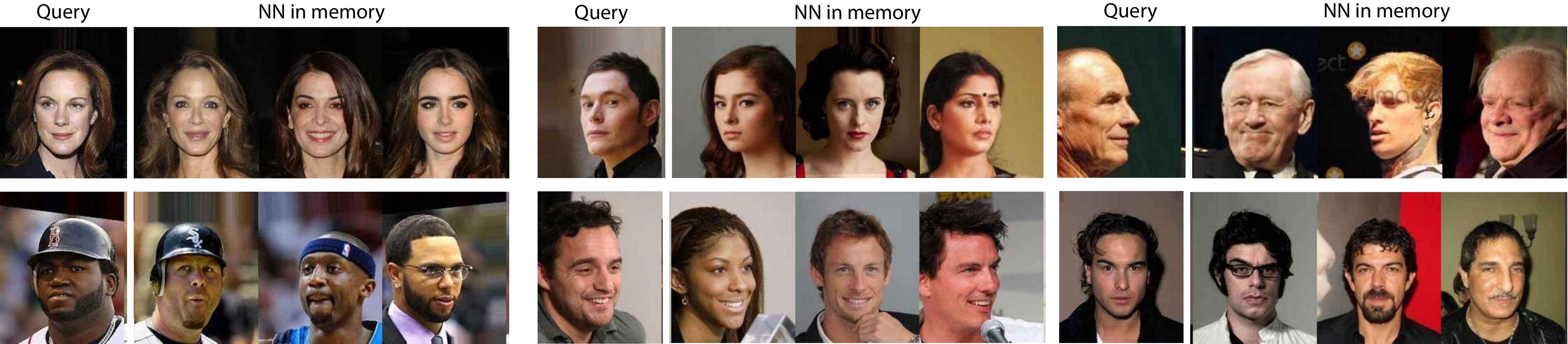}
    \caption{Despite never training on faces, the Shaders KML Mixup model can match for appearance and facial expressions on CelebA.}
    \label{fig:celeba-similarity}
\end{figure*}

\begin{table*}[t]
    %\vspace{-2cm}
    \centering
    \begin{adjustbox}{max width=\linewidth, max height=.15\textheight}
% \usepackage{booktabs}
% \usepackage{multirow}
\begin{tabular}{@{}ll|rrrrrrrrrr@{}}
Data type                  & Dataset       & Path           & Derma          & OCT            & Pneum          & Breast         & Blood          & Tissue         & OrganA         & OrganC         & OrganS         \\ \midrule
\multicolumn{2}{l|}{Best from original paper} & 91.1           & 76.8           & 77.6           & 94.6           & 86.3           & 96.6           & \textbf{70.3}  & \textbf{95.1}  & 92.0           & 81.3           \\ \midrule
\multirow{2}{*}{Realistic}    & ImageNet      & 98.34          & 82.05          & \textbf{95.11} & 95.23          & 91.03          & 96.79          & 57.00          & \textbf{94.33} & \textbf{93.81} & \textbf{85.89} \\
                              & Places        & 98.21          & 80.96          & 94.52          & 96.18          & 91.03          & 95.85          & 54.83          & 90.09          & 92.56          & 82.26          \\ \midrule
\multirow{5}{*}{Procedural}   & S. KML Mixup  & 98.60          & 81.36          & 92.85          & \textbf{96.56} & \textbf{93.59} & 95.09          & 55.62          & 83.76          & 81.48          & 74.43          \\
                              & S. KML        & \textbf{99.30} & \textbf{82.15} & 92.12          & \textbf{96.56} & 91.03          & 96.26          & 58.02          & 85.07          & 85.37          & 81.57          \\
                              & S. Mixup      & 98.88          & 81.56          & 89.14          & \textbf{96.56} & 92.31          & 96.73          & 53.27          & 81.96          & 82.53          & 74.10          \\
                              & Shaders       & 98.96          & 81.26          & 92.09          & 95.61          & 92.31          & \textbf{97.72} & \textbf{58.74} & 86.83          & 86.04          & 81.69          \\
                              & Stylegan      & 98.05          & 77.67          & 85.04          & 95.04          & 87.18          & 91.59          & 53.10          & 76.81          & 77.30          & 76.06          \\ \midrule
White-box                     & Random init.  & 73.05          & 67.50          & 66.39          & 84.54          & 83.33          & 85.51          & 44.27          & 75.07          & 68.94          & 60.93         
\end{tabular}
    \end{adjustbox}
    \caption{KNN classification accuracy on the MedMNIST datasets. Procedural models match or exceed the best result from the original paper \cite{yang_medmnist_2022} (a normally trained ResNet) in 7/10 datasets.}
    \label{tab:medical}
    %\vspace{-0.9cm}
\end{table*}

\textbf{Sensitive data} is information that legally or ethically needs to be handled with high standards of care and control, such as facial identity or medical data. In this scenario, directly training on the data is often not acceptable; procedural models with memory thus offer an elegant solution. \cref{fig:celeba-similarity} shows CelebA \cite{liu_deep_2015} query images and nearest neighbours according to the S. KML Mixup model. The latter effectively matches appearance and facial expressions despite never training on faces. \cref{tab:medical} shows classification accuracy on the MedMNIST datasets \cite{yang_medmnist_2021}, where we see that the procedural models match or exceed the best result from the paper in 7 out of the 10 datasets studied, and obtain good performance otherwise. While training on generic real data suppresses medical privacy risks, it still has identity privacy and bias issues.

\section{Model size scaling}
\label{sec:scaling-laws}

\begin{figure}[t]
    \centering
    \includegraphics[width=\linewidth, height=.2\textheight, keepaspectratio=True]{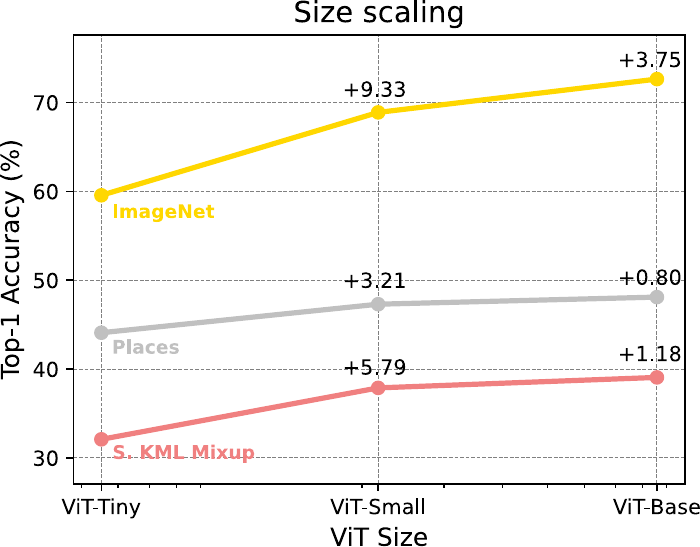}
    \caption{KNN classification accuracy on ImageNet-1K as Vision Transformer (ViT) size is scaled. Procedural models dont overfit as size increases; higher capacity yields higher performance.}
    \label{fig:scaling-size}
    %\vspace{-1cm}
\end{figure} 

\cref{fig:scaling-size} plots classification accuracy on ImageNet-1K versus model size. We observe that models do not overfit to procedural data as capacity increases, suggesting that with larger models performance increases. It also helps support the idea that procedural data models learn vision skills with good generalization.

\begin{figure*}[t]
    \centering
    \includegraphics[width=\linewidth]{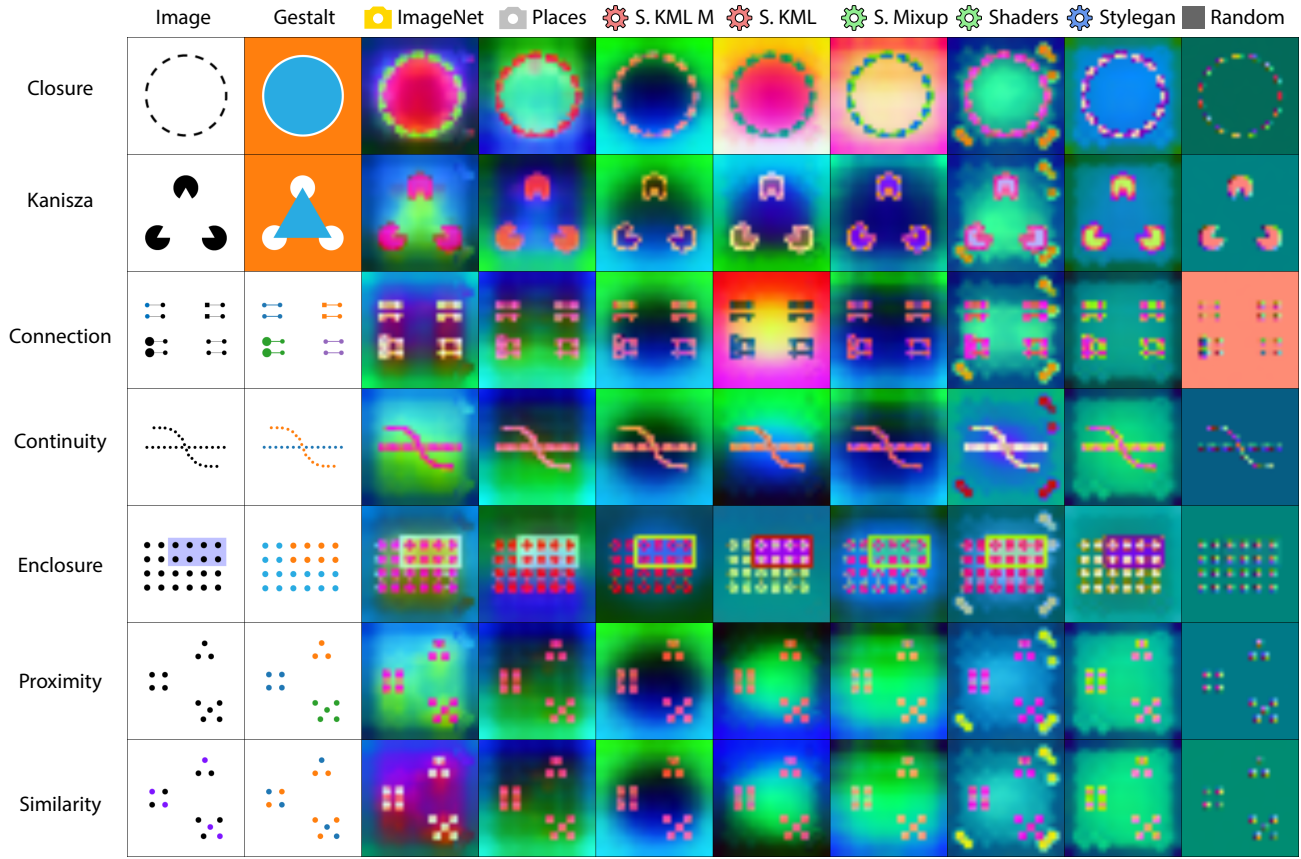}
    \caption{Feature PCAs for gestalt images. All vision models, real or procedural, do not seem to have gestalt perception.}
    \label{fig:gestalt}
\end{figure*}

\begin{table*}[t]
    \centering
    \begin{adjustbox}{max width=\linewidth, max height=\textheight}% Please add the following required packages to your document preamble:
\begin{tabular}{@{}cl|rrrrrrr@{}}
\multicolumn{1}{l}{Data type} & Dataset      & \multicolumn{1}{l}{Closure} & \multicolumn{1}{l}{Kanisza} & \multicolumn{1}{l}{Connection} & \multicolumn{1}{l}{Continuity} & \multicolumn{1}{l}{Enclosure} & \multicolumn{1}{l}{Proximity} & \multicolumn{1}{l}{Similarity} \\ \midrule
\multirow{2}{*}{Realistic}    & ImageNet     & \textbf{53.1}               & \textbf{21.7}               & 6.0                            & 6.3                            & 15.8                          & 24.4                          & 4.9                            \\
                              & Places       & 29.0                        & 12.7                        & 11.7                           & 17.3                           & 25.6                          & 47.1                          & 3.1                            \\ \midrule
\multirow{5}{*}{Procedural}   & S. KML Mixup & 36.1                        & 10.4                        & 45.1                           & 9.3                            & 31.9                          & \textbf{65.1}                 & 1.3                            \\
                              & S. KML       & 34.5                        & 14.6                        & \textbf{52.7}                  & 17.7                           & \textbf{64.8}                 & 52.0                          & 21.5                           \\
                              & S. Mixup     & 22.8                        & 15.4                        & 15.0                           & 31.1                           & 1.1                           & 30.2                          & 4.3                            \\
                              & Shaders      & 36.6                        & 15.8                        & 8.7                            & 40.0                           & 26.0                          & 12.1                          & 50.9                           \\
                              & Stylegan     & 25.4                        & 12.2                        & 20.4                           & \textbf{46.3}                  & 21.7                          & 7.9                           & 64.8                           \\ \midrule
\multirow{2}{*}{White-box}    & Random       & 0.9                         & 1.0                         & 17.6                           & 13.3                           & 2.8                           & 4.3                           & 21.9                           \\
                              & RGB          & 1.2                         & 0.0                         & 13.4                           & 28.5                           & 4.7                           & 28.9                          & \textbf{73.4}                 
\end{tabular}\end{adjustbox}
    \caption{$R^2$ of PCA features and gestalt segmentations.}
    \label{tab:gestalt}
\end{table*}

\section{Gestalt analysis}
In this section we evaluate wether unsupervised vision models, real or procedural, have an important characteristic of human perception: the gestalt. Introduced in 1920 by german psychologists Max Wertheimer, Kurt Koffka, and Wolfgang Kohler \cite{wertheimer_laws_1923,kohler_gestalt_1992,koffka_k_principles_1935}, gestalt theory posits that humans interpret scenes by abscribing a unified meaning to individual visual elements \cite{interaction_design_foundation_what_2016}. Graphic designers have since used the gestalt principles to great effect, making attractive designs with minimal ink by exploiting how humans will perceive them. The most important principles are the following: 
\begin{enumerate}
    \item Closure: perceiving dashed boundaries as continuous, and separating their interior and exterior.
    \item Kanisza: perceiving complete contours from partial contours.
    \item Connection: perceiving connected elements as a whole.
    \item Continuity: perceiving individual elements arranged continuously as part of a whole.
    \item Enclosure: perceiving elements contained inside a figure as different from equal elements outside of the figure
    \item Proximity: perceiving spatially close elements as part of a whole.
    \item Similarity: perceiving visually similar elements as part of a whole, even if far away (in particular, similarity supersedes proximity).
\end{enumerate}
\cref{fig:gestalt} shows exemplar images of the principles, their gestalt perception, and the PCA features of the images for real and procedural vision models. We qualitatively observe that with the exception of Closure, overall all the models (both real and procedural) do not have gestalt perception, which is numerically confirmed by the results in \cref{tab:gestalt}. This suggests that while vision models may be able to recognize objects and shapes, they are not yet perceptually aligned with humans.

\section{Storage, computation, and accuracy trade-offs of memory-based versus parametric classifier-based approaches in practical deployments}

Choosing between a memory-based approach and a parametric classifier-based parametric approach in practice involves a number of trade-offs. 

{\bf Training cost: } First and foremost is the computational cost of training. Training the linear classifier required 4 GPU-days on an 8-V100 node, while computing the embeddings needed for KNN classification required just 1.5 GPU-hours and is doable on a single GPU. The ratio is $\sim$64X. If the set of classes (the "world") is static, then the initial training is a one-time cost that may be acceptable. However, the real world is anything but static, and so re-training costs can quickly spiral.

{\bf Inference cost:} On the original research code, the inference costs of the KNN classifier were about double of the linear classifier's (1min52s vs 4min for the entire ImageNet validation set of 50k). However, dedicated efficient nearest neighbour search libraries such as faiss \cite{faiss,faiss-gpu} can make search much more efficient in a production setting i.e. with queries within a 10M database taking $<$0.03ms.

{\bf Memory cost:} The storage requirements is where the memory-based approach is most demanding in comparison with the classifier approach, as the former scales with the number of examples while the latter with the number of classes. On ImageNet, this ratio is about 1000X. However, modern storage technology is incredibly cheap especially compared to GPUs. Storing the entire ImageNet embeddings (384 floats x 1.3M examples) would require $\sim$2GiB, and 1000GiB SSDs may be purchased online for $\sim$100USD. In contrast for GPUs, a single V100 has MSRP around 10,000USD while a single H100 has MSRP 30,000 USD, which are needed to train the models. Essentially, storage costs are almost negligible compared to training and inference costs.

\section{Conclusion}
Deep learning is an extremely powerful framework, capable of learning with very little explicit structure by optimizing weights on data. However, representing knowledge via weights has some important drawbacks. In particular, it is very difficult to add, remove, and edit knowledge once its been put into weights. As models scale, retraining or fine-tuning becomes more and more costly, especially in a future where data is licensed instead of bought, as most digital goods are right now. Correcting mistakes or eliminating outdated data as our understanding of reality grows is also of high interest. Privacy-wise, regulations currently limit many AI tools in the EU. To this end, prior work proposed an explicit separation between how knowledge is represented, the feature embeddings, and what knowledge is stored, the visual memory. Taking the form of a database of instances, editing knowledge in the visual memory is as simple as adding or dropping data. However, a problem remained: the feature embeddings themselves were generated through weights trained on real-world data, so perception and knowledge remained entangled. In this work, we proposed training the embeddings with procedural data. Unlike real-world data, procedural data is non-realistic and generated via simple code, and thus has a degree of separation from the real word. In particular, it is much less exposed to the privacy or bias risks that necessitate unlearning. Combining procedural embeddings and visual memory results in a system where \emph{all} real world data can be flexibly added, removed, and provably evaluated for privacy, while approximating the performance of standard methods.
\vspace{-0.15cm}

\section*{Acknowledgements} This work was supported by the Defense Science and Technology
Agency, Singapore. Additionally, Adrian Rodriguez-Muñoz was supported by the LaCaixa fellowship (LCF/BQ/EU22/11930084).

\section*{Impact Statement}
Our work contributes towards AI models that are more transparent, controllable, and applicable in domains with strict data regulations. Moreover, minimizing fine-tuning or retraining significantly reduces power demands and allows remaining power to be allocated in more useful ways.


% \clearpage
\bibliography{ProceduralTraining}
\bibliographystyle{icml2025}

%%%%%%%%%%%%%%%%%%%%%%%%%%%%%%%%%%%%%%%%%%%%%%%%%%%%%%%%%%%%%%%%%%%%%%%%%%%%%%%
%%%%%%%%%%%%%%%%%%%%%%%%%%%%%%%%%%%%%%%%%%%%%%%%%%%%%%%%%%%%%%%%%%%%%%%%%%%%%%%
% APPENDIX
%%%%%%%%%%%%%%%%%%%%%%%%%%%%%%%%%%%%%%%%%%%%%%%%%%%%%%%%%%%%%%%%%%%%%%%%%%%%%%%
%%%%%%%%%%%%%%%%%%%%%%%%%%%%%%%%%%%%%%%%%%%%%%%%%%%%%%%%%%%%%%%%%%%%%%%%%%%%%%%
\newpage
\appendix
\onecolumn
% AC recommends acceptance and strongly encourages the authors to include an analysis or discussion of the trade-offs between storage, computation, and accuracy in the final version to further strengthen the practical relevance of the work.

\section{Training details}

We trained a vision transformers (Small ViT) \cite{dosovitskiy_image_2021} for each dataset (ImageNet, Places, Shaders KML Mixup, Shaders KML, Shaders Mixup, Shaders, and Stylegan), using the recipe and architecture of the original DINO paper \cite{caron_emerging_2021}. In particular, we used the optimal hyperparameters for the model trained on ImageNet for all models, rather than hyper-optimizing for performance on each specific dataset. This results in a much more rigorous evaluation, as the optimal ImageNet hyperparameters are more likely to be bad than good for procedural non-realistic data. These hyperparameters are: learning rate 1e-3, batch size 512, optimizer AdamW, num epochs 100, and DINO head out dim 65536. All models are then used without any fine-tuning to obtain all the results, including Figure 5, Table 1, Figures 9 and 10, and Tables 2 and 3.

\section{Additional results}

\subsection{Benchmark saturation on NIGHTS}

In \cref{fig:nights-hvs-perf} it visually appears that Shaders-based procedural models are all quite close in performance to each other and to the real data Places model. To quantitatively test this, we performed a z-test and determined that Places, S. KML, and Shaders are all equivalent to the 5\% level. This supports the finding that procedural models have reached the level of real models this benchmark. For the z-test, we used the average NIGHTS results and number of samples in the val dataset (1720). We also include standard deviations of the mean for reference in \cref{tab:nights_val-std}.

\begin{table}[h]
\centering
\begin{tabular}{@{}l|l|r@{}}
\multicolumn{1}{c}{Dataset Type} & \multicolumn{1}{|c|}{Dataset} & \multicolumn{1}{c}{NIGHTS-Val} \\ \midrule
Target                          & ImageNet                     & 0.8733 $\pm$ 0.0080                       \\ \midrule
Real                       & Places                       & 0.8331 $\pm$ 0.0090                       \\ \midrule
\multirow{5}{*}{Procedural}     & S. KML Mixup                 & 0.8105 $\pm$ 0.0095             \\
                                & S. KML                       & 0.8244 $\pm$ 0.0092                       \\
                                & S. Mixup                     & 0.8110 $\pm$ 0.0094                       \\
                                & Shaders                      & 0.8169 $\pm$ 0.0093                       \\
                                & Stylegan                     & 0.7605 $\pm$ 0.0103                       \\
\end{tabular}
\caption{NIGHTS-Val performance with standard deviations.}
\label{tab:nights_val-std}
\end{table}

% Linear decoding and gradcam visualizations
\subsection{Linear decoding performance}

% \usepackage{booktabs}
% \usepackage{multirow}
% \usepackage[normalem]{ulem}
\begin{table}[h]
\centering
\begin{tabular}{@{}l|l|r|r@{}}
\multicolumn{1}{c}{Dataset Type} & \multicolumn{1}{c|}{Dataset} & \multicolumn{1}{|c|}{KNN} & \multicolumn{1}{c}{Linear} \\ \midrule
Target                          & ImageNet                     & 68.9                      & 73.6                       \\ \midrule
Real                       & Places                       & 47.3                      & 59.6                       \\ \midrule
\multirow{5}{*}{Procedural}     & S. KML Mixup                 & \textbf{37.9}            & \textbf{47.3}             \\
                                & S. KML                       & 35.4                      & 47.1                       \\
                                & S. Mixup                     & 35.4                      & 44.8                       \\
                                & Shaders                      & 30.7                      & 43.1                       \\
                                & Stylegan                     & 13.7                      & 26.4                       \\
\end{tabular}
\vspace{0.5cm}
\caption{Top-1 accuracy results for linear decoding on ImageNet-1K. With linear decoding, S. KML beats S. Mixup by 2.2\%, despite the KNN performances being equivalent. Moreover, the gains from adding Mixup to both Shaders and S. KML are much smaller. This suggests that Mixup mainly reduces bad features, which can also be pruned by the decoder, while KML yields either better or a greater amount of useful features. The two approaches are complementary, which is why Shaders KML Mixup obtains the strongest performance overall.}
\label{tab:linear-decoding}
\end{table}

\subsection{Linear decoding GradCam}
\begin{figure}[H]
    \centering
    \includegraphics[width=\linewidth,height=\textheight,keepaspectratio]{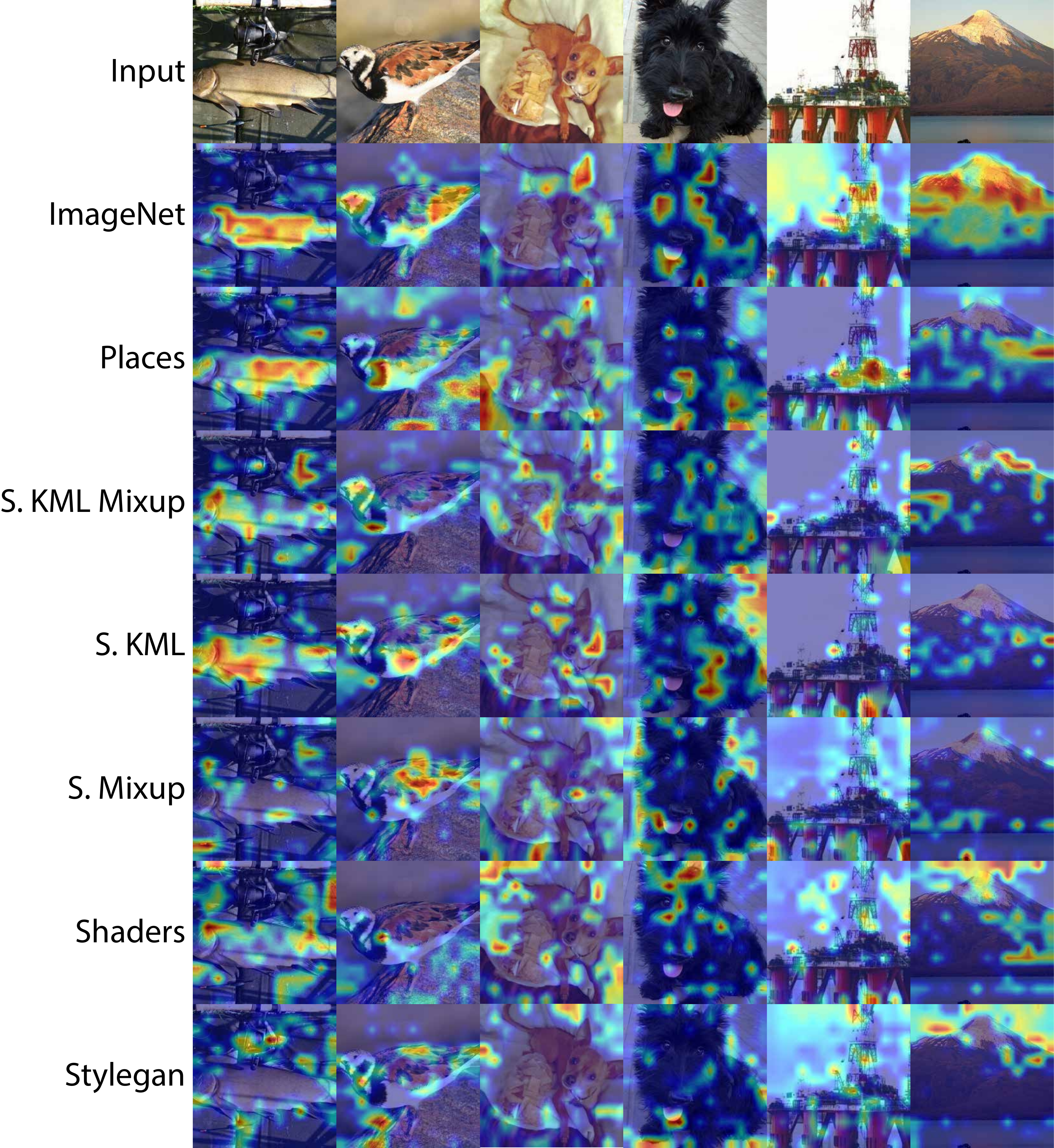}
    \caption{GradCam \cite{gradcam,gradcam-code} visualization for linear decoding on random images from ImageNet for each of the models.}
    \label{fig:gradcam}
\end{figure}

\subsection{Hard-label unsupervised segmentation with K-Means}

\begin{figure}
    \centering
    \includegraphics[width=.9\linewidth]{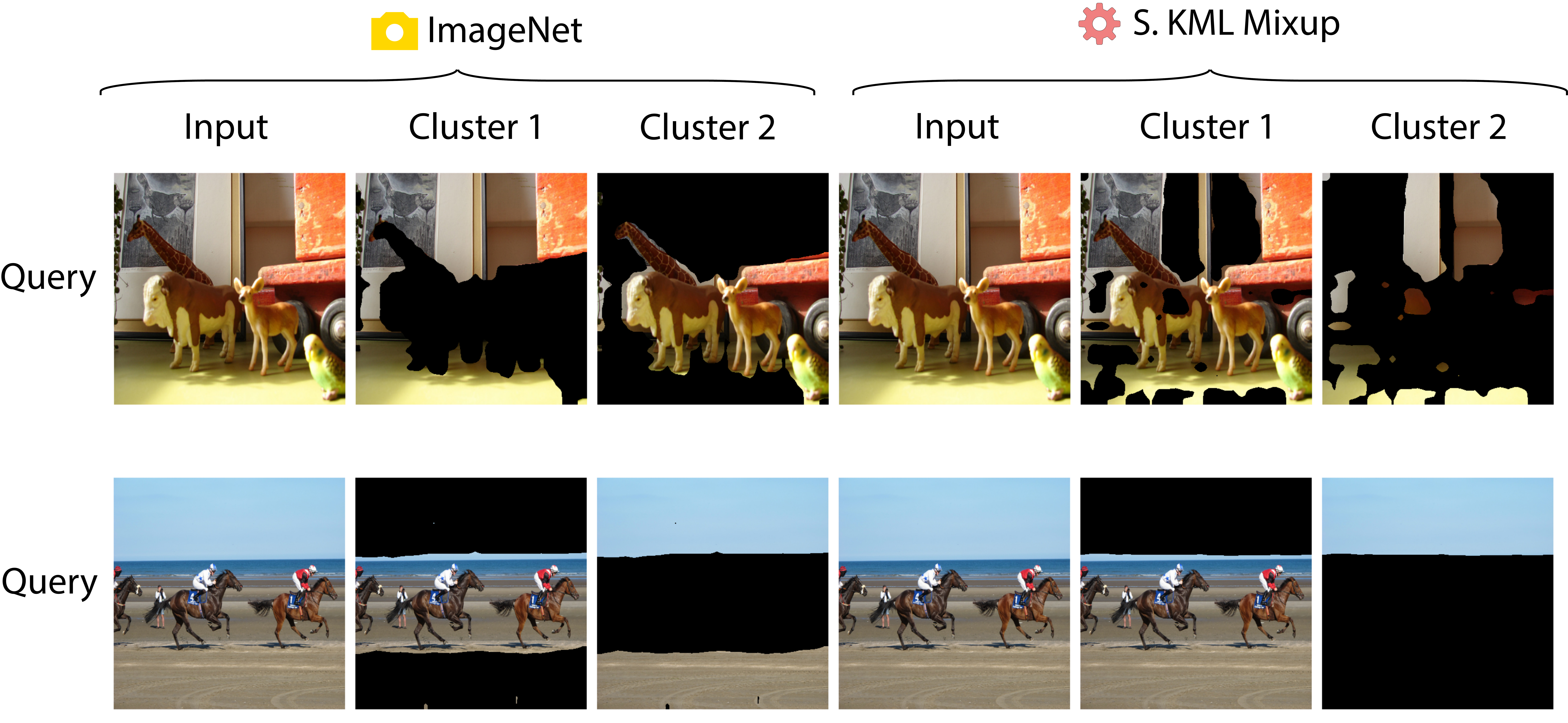}
    \caption{Hard-level unsupervised segmentation with K-Means on COCO. The procedural S. KML Mixup model makes visually sensible clusters, but that do not reflect real-world objects, unlike the ImageNet model. This results in incorrect class nearest neighbours, as seen in \cref{fig:hard-label-12}}
    \label{fig:hard-label-9}
\end{figure}

\begin{figure}
    \centering
    \includegraphics[width=.9\linewidth]{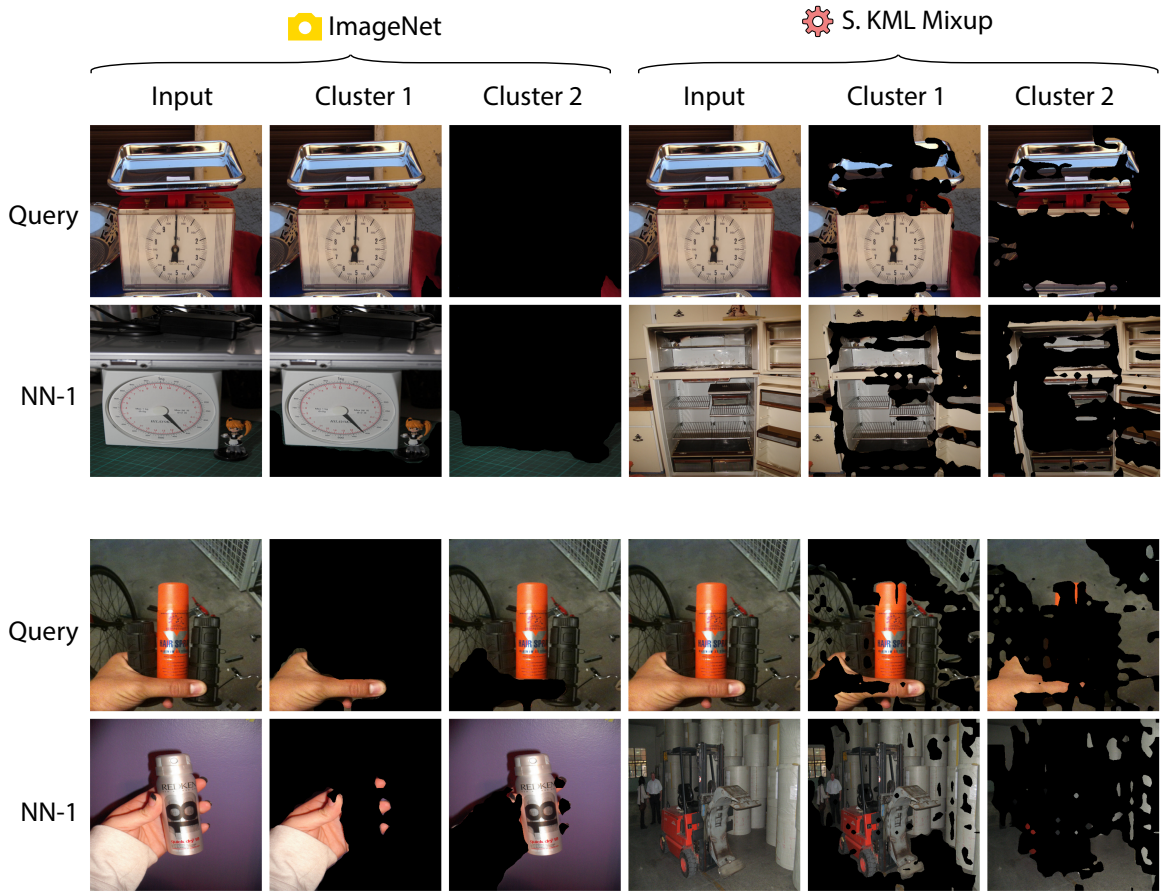}
    \caption{Hard-level unsupervised segmentation with K-Means on ImageNet of a query image and is Nearest Neighbour (NN-1). The procedural S. KML Mixup model makes visually sensible clusters, but that do not reflect real-world objects, unlike the ImageNet model. The clusters of the query and NN-1 images visually resemble each other, but since the visual resemblance is not aligned with real world objects, the choices are incorrect for classification.}
    \label{fig:hard-label-12}
\end{figure}

\end{document}